\documentclass[lettersize,journal]{IEEEtran}
\usepackage{amsmath,amsfonts}
\usepackage{algorithmic}
\usepackage{algorithm}
\usepackage{array}
\usepackage[caption=false,font=normalsize,labelfont=sf,textfont=sf]{subfig}
\usepackage{textcomp}
\usepackage{stfloats}
\usepackage{url}
\usepackage{verbatim}
\usepackage{graphicx}
\usepackage{cite}
\begin{document}

\title{The Quiet Eye Phenomenon in Minimally Invasive Surgery}

\author{Alaa Eldin Abdelaal$^{1}$~\IEEEmembership{Member, IEEE}, Rachelle Van Rumpt$^{2}$, Sayem Nazmuz Zaman$^{3}$, Irene Tong$^{2}$, Anthony Jarc$^{4}$, Gary L. Gallia$^{5}$, Masaru Ishii$^{6}$, Gregory~D.~Hager$^{7}$~\IEEEmembership{Fellow, IEEE}, and Septimiu E. Salcudean$^{1}$~\IEEEmembership{Fellow, IEEE}
        % <-this % stops a space
\thanks{Manuscript received September 6, 2023. This work was supported in part by the Natural Sciences and Engineering Research Council of Canada (Discovery Grant), in part by the Canada Foundation for Innovation (infrastructure and operating funds), in part by Intuitive Surgical (equipment donation), in part by an Intuitive Foundation Clinical Research Grant, in part by the C.A. Laszlo Chair in Biomedical Engineering held by Prof. Salcudean, and in part by the Vanier Canada Graduate Scholarship and a Michael Smith Foreign Study Supplement both held by Alaa Eldin Abdelaal.}% <-this % stops a space
\thanks{$^{1}$A. E. Abdelaal is with the Mechanical Engineering Department, Stanford University, Stanford, USA. 
{\tt\small (email: abdelaal@stanford.edu)}}
\thanks{$^{2}$ R. Van Rumpt, I. Tong and S. E. Salcudean are with the Electrical and Computer Engineering Department, University of British Columbia, 2332 Main Mall, Vancouver, BC Canada.
        }%
\thanks{$^{3}$S. N. Zaman is with the Department of Physics and Astronomy, University of British Columbia, 325 - 6224 Agricultural Road, Vancouver, BC Canada.}
\thanks{$^{4}$A. Jarc is with Intuitive Surgical, Atlanta, GA, USA.}
\thanks{$^{5}$G. L. Gallia is with Department of Neurosurgery, Johns Hopkins University School of Medicine, Baltimore, MD, USA.}
\thanks{$^{6}$M. Ishii is with  Department of Otolaryngology – Head and Neck Surgery, Johns Hopkins University School of Medicine, Baltimore, MD, USA}          
\thanks{$^{7}$G. D. Hager is with Department of Computer Science, Johns Hopkins University, Baltimore, MD 21218, USA.}
}

%\markboth{IEEE Transactions on Medical Robotics and Bionics}%
%{Abdelaal \MakeLowercase{\textit{et al.}}: The Quiet Eye Phenomenon in Minimally Invasive Surgery}


\maketitle

\begin{abstract}
In this paper, we report our discovery of a gaze behavior called Quiet Eye (QE) in minimally invasive surgery. The QE behavior has been extensively studied in sports training and has been associated with higher level of expertise in multiple sports. We investigated the QE behavior in two independently collected data sets of surgeons performing tasks in a sinus surgery setting and a robotic surgery setting, respectively. Our results show that the QE behavior is more likely to occur in successful task executions and in performances of surgeons of high level of expertise. These results open the door to use the QE behavior in both training and skill assessment in minimally invasive surgery.\end{abstract}


\section{Introduction}
\label{sec:introduction}
\IEEEPARstart{S}{tudying} how experts see their workspace when performing a task is one of the main research topics in motor skill learning. In particular, researchers in this area are interested in investigating where experts look in their workspace, when they look at critical cues, for how long, and how this behavior is coordinated with the body/hand motions of these experts. Answering these questions can lead to interesting insights that can then be used to train novices. 

Indeed, gaze-based training methods have shown great success in many fields including sports~\cite{causer2010quiet}, law enforcement~\cite{vickers2012performing} and the military~\cite{ward2008training}. The general idea of these methods is to train novices to adopt the gaze behaviors of experts. The essence of these methods is to teach novices where and when to look at locations of interest and, for how long. The underlying argument is that the motor skills of novices can be improved if they adopt the “visual behavior” of experts. 

One promising gaze-based training method is called quiet eye (QE) training. The term “quiet eye” refers to the final fixation right before a performer carries out a critical movement to execute a task~\cite{vickers2009advances}. The period of this final fixation is called the QE period. Research in sports shows that a longer QE period is associated with a higher level of expertise~\cite{williams2002quiet}. This QE behavior has been discovered in experts' performances in targeting tasks in sports such as basketball free throws~\cite{vickers1996visual} and golf putting~\cite{vickers1992gaze}. One theory explaining such behavior states that this long QE period is the period needed to change the neural connections in the performer’s brain to absorb and process the important visual information and plan the critical movement of the task before actually carrying it out~\cite{mann2011quiet}.

As many surgical tasks can be considered as targeting tasks, QE training has the potential to be applied in surgical training, inspired by the strong evidence of its effectiveness in other domains. Studies in this area have been limited and are focused solely on open surgery settings. In this work, we investigate the QE phenomenon in the context of minimally invasive surgery (MIS). In particular, the contributions of this work are: 
\begin{itemize}
	\item To investigate the existence of QE behavior in surgeons’ performances in two minimally-invasive surgery (MIS) settings. Our investigation is conducted on two independently collected data sets, one in a sinus surgery setting and the other in a robotic surgery setting using the da Vinci Surgical robot. %add trademark sign?
	\item To observe and report the changes in the QE behavior in successful and unsuccessful tasks. %or performances?
	\item To observe and report the changes in the QE behavior between highly experienced and less experienced surgeons.
\end{itemize}
These contributions provide the foundation to understand the QE behavior in MIS and open the door for further research on the effect of applying QE training in this setting. 

\section{Related Work}
\label{sec:Related Work}

Researchers in sports training have extensively studied the eye gaze of experts as a proxy of how they allocate their attention. To achieve that, they follow what is called the vision-in-action paradigm for data collection~\cite{vickers1996visual}, where they collect the gaze data of experts as they actually perform the task. In addition, they also use cameras to capture synchronized views of the body of the experts as well as views of what they see in their workspace. The analysis of the collected data has been used to find interesting gaze patterns in experts and study how such patterns differ from novices. Such an approach has been successfully applied in many sports such as basketball~\cite{ripoll1986stabilization}, pistol shooting~\cite{ripoll1985analysis}, badminton~\cite{ripoll1988does} and golf~\cite{vickers1992gaze}.

A phenomenon called Quiet Eye (QE) was discovered by Joan Vickers when analyzing the gaze behavior of experts in basketball~\cite{vickers1996visual} and it was identified and associated with experts’ performances in many other sports later on as in~\cite{williams2002quiet, panchuk2017quiet, sun2016quiet}. QE is defined as the final fixation (or tracking gaze) right before a performer carries out a critical movement to execute a task~\cite{vickers2009advances}. The period of this behavior is called the QE period. 

Thus far, the QE phenomenon has been confirmed in three main task categories: targeting tasks~\cite{vine2014quiet} (such as basketball free throws and billiards), interceptive timing tasks~\cite{sun2016quiet} (such as responding to the serve in volleyball and table tennis) and tactical tasks~\cite{williams1998visual} (such as soccer defence and offence). Adopting the QE behavior of experts has been shown to be effective in training novices in these three task categories.

%Sports research shows that QE behavior occurs in experts’ performance in three major
%categories of motor tasks: (i) Targeting tasks such as basketball free throws and golf putting, (ii)
%Interceptive timing tasks such as volleyball and table tennis serve reception, and (iii) Tactical
%tasks such as soccer offense and defence. We argue that many surgical tasks can be
%considered as targeting tasks (such as needle insertion, ablation, targeting in sinus surgery and
%so on). Therefore, we propose to investigate the existence of QE behavior in experts’
%performance in this category of tasks to validate the findings of sports research in minimally
%invasive surgery.

%QE training involves similar steps regardless of the task at hand [29]. First, the QE behavior is shown to trainees using a video of an expert performing the task with his/her eye gaze overlaid onto it. Trainees are then asked to practice performing the task while their gaze data is being recorded. After each trial, trainees are shown a video of their own performance with their eye gaze overlaid onto it as well as the video of the expert that they saw in the first step. The idea here is to let them see the differences between their own gaze behavior and the expert’s and then decide what to do in their next trial to eliminate these differences.

QE behavior can be identified by looking for the following measurable characteristics in the collected data~\cite{vickers2007perception}:
\begin{itemize}
	\item It is a fixation (or tracking gaze) on a specific target in the workspace which strictly means that the performer's gaze remains within 3 degrees of visual angle (or less) from the same target for at least 100 ms.
	\item The beginning of the QE period (which is also called QE onset) begins right before the critical movement of the task. For instance, if it is a targeting task, then the QE period begins right before its aiming part.
	\item The end of the QE period (or the QE offset) occurs when the fixation (or tracking gaze) deviates from the location/object of interest by more than 3 degrees of visual angle for longer than 100 ms.
\end{itemize}
%make et al italic, check the previous papers

The successful use of the QE in sports encouraged its exploration in surgical training. For example, Causer~et al.~\cite{causer2014quiet} conducted a user study to assess the efficiency of QE in open surgery setting in a knot tying task. In their study, the trainees were divided into two groups, a QE training group and a conventional training group. Their results show that the QE group performs better than the other one in terms of completion time and efficiency of the hand movements during the task. The results of another study~\cite{causer2014performing} also show that the QE training group performance is more robust to high levels of pressure and anxiety compared with the other group. Through analyzing the gaze behavior, the authors argue that QE training directs the trainees' gaze towards a single point of focus, which increases the effectiveness of their attention allocation to the most important information for the task at hand. This frees up some mental processing resources that enable QE trainees to overcome high anxiety levels.
% never studeid in MIS?

To the best of our knowledge, the QE phenomenon has neither been identified in surgeons' gaze behavior nor applied to surgical training in MIS settings. We argue that many surgical tasks fall under the targeting task category. Findings from sports research suggest that there is a very high chance of finding the QE phenomenon in surgeons' performances of tasks in this category in our case. In addition, according to Newell's Constraints-Led Model~\cite{newell1986constraints}, a change in the environment/setting is a significant factor to consider before studying (and later on applying) a motor skill learning phenomenon to other environments, which justifies the need to study the QE phenomenon in MIS settings (in contrast to open surgery settings as in previous work). 

In this work, we take the first step towards filling this gap by investigating the existence of the QE behavior in surgeons’ performances in two different MIS settings. In addition, we report our observations regarding how the QE behavior changes with successful/unsuccessful task executions, and with surgeons' experience level. We believe that by doing so, this work provides a deeper understanding of the QE behavior in MIS, which, in turn, can be beneficial in applications such as training and skill assessment in MIS. 

\section{Methods and Hypotheses}
\label{sec:Methods and Hypotheses}

\subsection{Hypotheses}
\label{subsec:Hypotheses}

In this work we have two hypotheses as follows:
\begin{itemize}
	\item Hypothesis 1: QE behavior occurs more frequently in successful tasks compared with unsuccessful ones. In addition, QE duration is longer in successful tasks than in unsuccessful ones.   
	\item Hypothesis 2: QE behavior occurs more frequently in performances from highly experienced surgeons compared with less experienced ones. In addition, QE duration is longer in the former case than the latter one.  
\end{itemize}
We verify the first hypothesis using a data set in a sinus surgery setting. The second hypothesis is verified using another data set in a robotic surgery setting using the da Vinci surgical robot. 

\subsection{Sinus Surgery Data Set}
\label{subsec:Sinus Surgery Data Set}
\begin{figure}[!t]
      \centering
			\captionsetup{justification=centering}
     \includegraphics[width=0.4\textwidth] {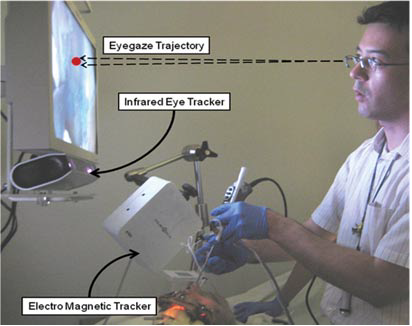}
      \caption{The experimental setup used for the sinus surgery data collection~\cite{ahmidi2012objective}. This figure is used with permission.} %check UBC guidlines on using figures with permission.
      \label{JHU_Data_Set_Collection}
   \end{figure}

%check for any use of the same text as in the MICCAI paper. -- done
The sinus surgery data set is a data set of a series of targeting tasks on a cadaver model~\cite{ahmidi2012objective}. The data set was collected from 20 surgeons at the annual resident endoscopic skull-base surgery course at Johns Hopkins University. The subjects were asked to visualize and touch nine endonasal structures using an endoscope and pointer. The experimental setup used for the data collection is shown in Fig.~\ref{JHU_Data_Set_Collection}. An electromagnetic tracker was used to track the positions of the pointer and endoscope, and an eye gaze tracker was used to collect the subject's eye gaze position on the video monitor every 0.02 seconds. Videos of what subjects saw were recorded directly using a video capture card situated in a data collection computer. Temporal synchronization between all the collected data units was achieved using time stamps assigned to each of these units.
% adjust the postion of this figure. In the PDF it is before its subsection starts. 

Each video in the data set represents a targeting task by one surgeon, aiming for a predefined target inside the sinus anatomy. Each video was assessed by three independent assessors. Each assessor gave each video a task score of either zero, if the task was not successful, or one if it was successful. This makes the data set a suitable candidate to investigate the QE behavior and how it changes with the successful execution of these targeting tasks.  

The video annotations were carried out manually based on the QE three measurable characteristics mentioned in Section~\ref{sec:Related Work}. In particular, we annotate the target location in the first frame in each video and track it (along with its surrounding region) in the subsequent frames using OpenCV. This gives us the 2D pixel location of the target in each frame. The raw data contains the 2D pixel location of the surgeon's eye gaze in each frame and its corresponding time stamp. Using this information, we compute all the fixations on the target location, within three degrees of visual angle, that last for at least 100 ms. The three degrees of visual angle are represented on each frame as a circle whose center is the target pixel location. The radius of this circle is computed by projecting the three degrees of visual angle from the surgeon's eye onto the screen. To compute this radius in pixels, we use the distance between the surgeon's eye and screen, the size of the screen and its resolution. All these values are known from the data collection process. A fixation is detected as long as the surgeon's eye gaze location is within this circle for at least 100 ms. Using time stamps, we also compute the duration of each detected fixation. To find out if QE occurs or not, we first manually identify the frame number right before the start of the aiming part of the targeting task, and its associated time stamp. Using this information, we conclude that QE occurs in a video if and only if a fixation starts at or before this time stamp and lasts after the start of the aiming part of the task. The QE duration is the same as the duration of this fixation. The data set was annotated by two independent annotators and the interclass correlation between them is 1 for QE detection and 0.9903 for QE duration. 
%Add a figure of the projection from the surgeon's eye to the screen. 

\begin{figure}
      \centering
     \includegraphics[width=0.4\textwidth] {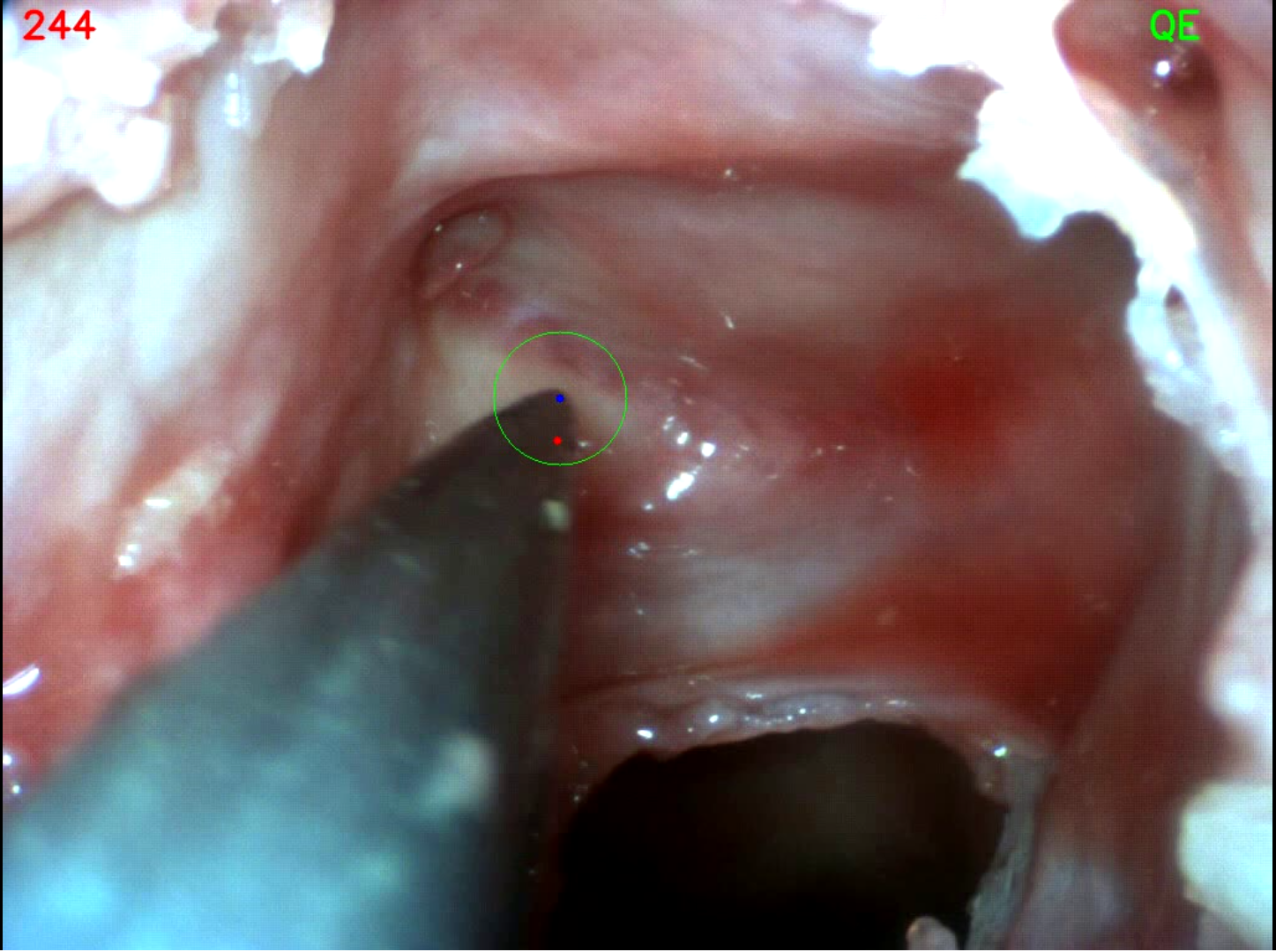}
      \caption{A sample annotated frame from one of the videos where QE occurred in the sinus surgery data set. The blue point represents the target location, the red point represents the surgeon’s eye gaze, and the green circle represents the area within $3$ degrees of visual angle around the target (according to the definition of QE). A QE green flag appears on the top right of the view when the QE occurred in the video.}
      \label{QE_JHU_Data_Set}
   \end{figure}

The data set has a total of 302 videos and we excluded 28 of them because they were difficult to annotate due to the unstable visual feedback caused by frequent camera movement. For the remaining 274 videos, we annotated each video so that we can learn two things:
\begin{itemize}
	\item Whether the QE exists in the video. In this case, the video was assigned a binary value (e.g., one if the QE occurs in the video and zero otherwise).
	\item The QE duration in milliseconds, only if QE exists in the video.
\end{itemize}
A sample annotated frame from one of the videos where QE occurred is shown in Fig.~\ref{QE_JHU_Data_Set}. %A sample annotated video from the data set can be accessed through this link: \url{https://drive.google.com/file/d/1HV_2TizDSNnp-Fy22aTsnwjjEbyJYfdx/view}
% we may need to thicken the circle and dots in this figure, they are not very clear for everyone. 
%Put this figure in the correct location, it is not right now. 

To verify the first hypothesis, we only considered the videos with unanimous task scores (either zero or one) from the three assessors. This represents a total of 187 videos from the entire data set.

\subsection{Robotic Surgery Data Set}
\label{subsec:QE_Robotic Surgery Data Set}
%adapt accrodngly + no mention of specific tasks and details about expereince level + cite Irene's thesis (done), check the our langue is different. Change the third sentence, it is very similar to Irene's (done)
The robotic surgery data set was collected from seven RAS surgeons operating on the da Vinci Si Surgical System~\cite{tong2017eye}. The surgeons performed porcine and cadaver exercises. During each exercise, the estimated eye gaze for each surgeon's left and right eye was collected using EyeTech VT2 mini eye tracker (EyeTech Digital Systems, Mesa, AZ), that was placed within the da Vinci Si system. The surgeon's point of gaze was then calculated as the mean of the left and right eye locations. In addition, the instrument kinematics, system events, and endoscope videos were collected directly from the da Vinci platform, and were synchronized with the eye gaze data using the associated time stamps.   
%Add figure from the annotated suturing vidoes (done) + clearly mention the trageting component of the suturing task (done)

\begin{figure}[!t]
      \centering
     \includegraphics[width=0.4\textwidth] {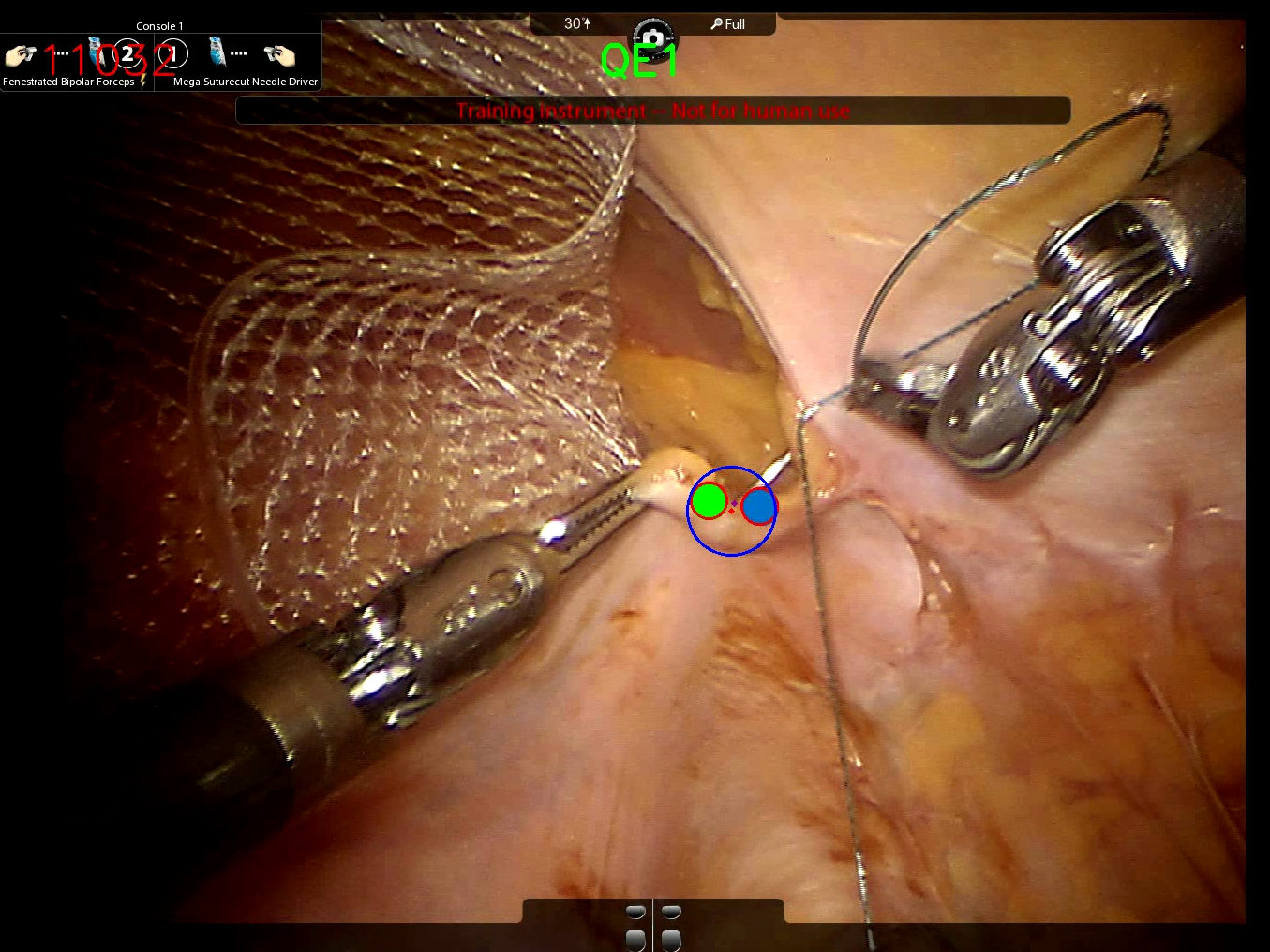}
      \caption{A sample annotated frame from one of the videos where QE occurred in the robotic surgery data set. The blue point represents estimated gaze for the surgeon's right eye and the green point represents estimated gaze for the surgeon’s left eye. The blue circle represents the area within 3 degrees of visual angle around the target (according to the definition of QE). A QE green flag appears on the top middle of the view when the QE occurred in the video.}
      \label{QE_IS_Data_Set}
   \end{figure}

Since we are interested in targeting tasks, we annotated the videos of all the suturing exercises in the data set. This is because the suturing task has a clear targeting component whenever the surgeon inserts the needle into the tissue. The annotation was performed manually, similar to the annotation of the sinus surgery data set. A sample annotated frame from one of the videos where QE occurs is shown in Fig.~\ref{QE_IS_Data_Set}

%Put this figure in the correct location, it is not right now. 
%put range of years of expereince? + say that other surgeons did tasks that are not clearly targeting tasks (done). 

In total, we had 33 videos of suturing exercises performed by two surgeons. The first surgeon has approximately 20 years of surgical experience and the second surgeon has less than 5 years of surgical experience. The remaining surgeons in the data set did not perform any suturing exercises. The results from these videos are used to verify our second hypothesis. 

%how we end up using two surgeons + their expereince level (done) + their tasks and no of videos

\subsection{Performance Metrics}
\label{subsec:QE_Performance Metrics}

We considered in our investigation two metrics: (i) the percentage of the videos that have QE for each considered task score/experience level, and (ii) the QE duration in these videos. %remove the average and std? 

\section{Results}
\label{sec:QE_Results}
%have two subsections for each data set? 

\subsection{Sinus Surgery Data Set Results}
\label{subsec:Sinus Surgery Data Set Results}

\begin{figure}[!t]
      \centering
			\captionsetup{justification=centering}
     \includegraphics[width=0.45\textwidth] {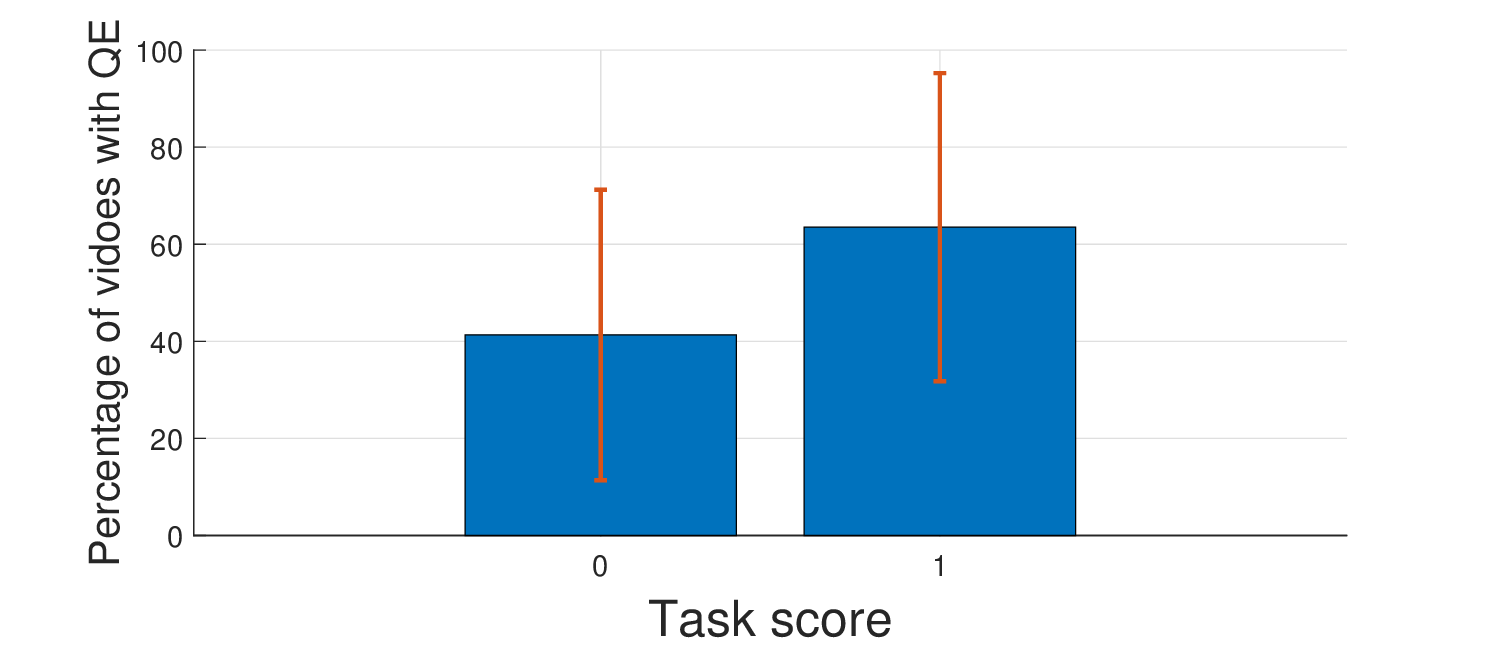}
      \caption{The percentage of the videos that have QE for each considered task score.}
      \label{QE_Percentage_JHU}
   \end{figure}

\begin{figure}[!t]
      \centering
     \includegraphics[width=0.45\textwidth] {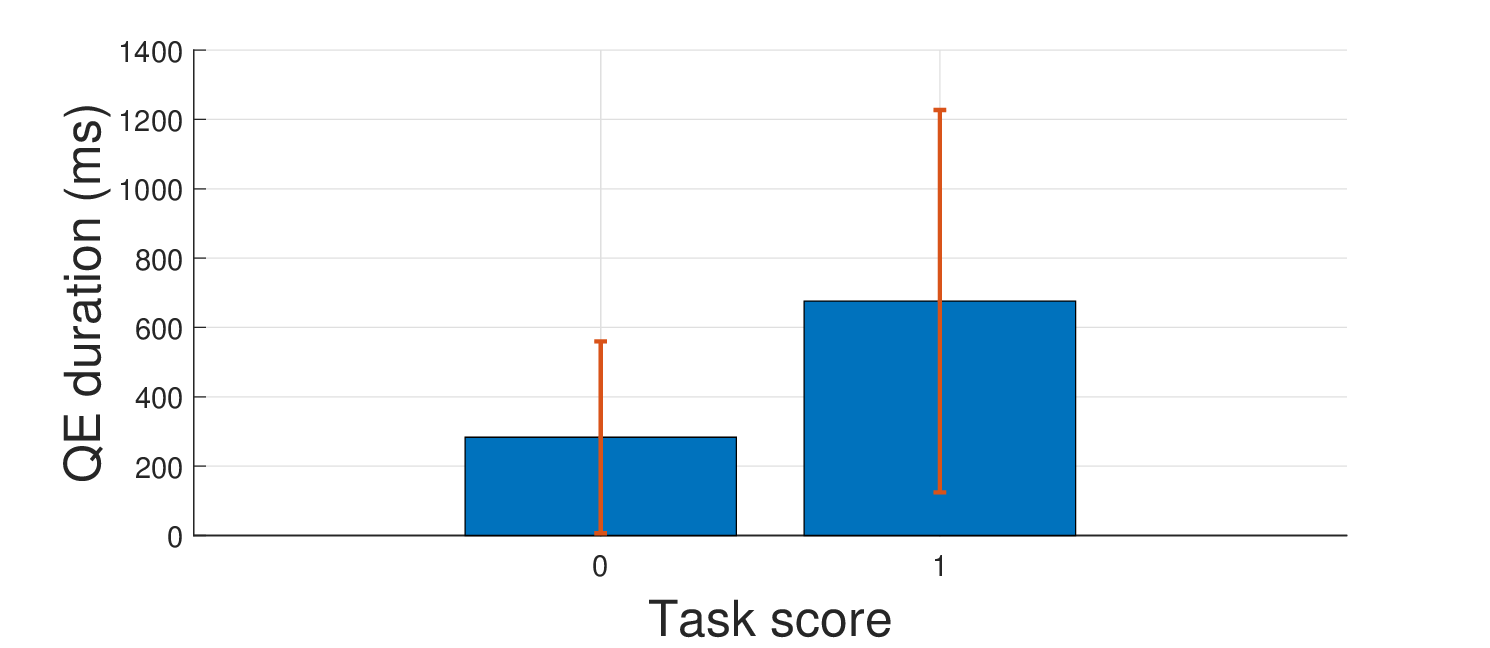}
      \caption{The QE duration for each considered task score.}
      \label{QE_duration_JHU}
   \end{figure}

The results of the sinus surgery data set, as shown in Fig.~\ref{QE_Percentage_JHU} and Fig.~\ref{QE_duration_JHU}, are used to verify our first hypothesis. Results show that videos with a unanimous task score of one have on average 22\% more videos with QE compared with videos with a task score of zero as shown in Fig.~\ref{QE_Percentage_JHU}. Furthermore, the QE duration in videos with a task score of one is on average 2.3 times longer than the QE duration in videos with a task score of zero as shown in Fig.~\ref{QE_duration_JHU}.

We perform hypothesis testing using a Mann–Whitney U test on the above results. The two results were statistically significant, with \textit{p}~$<$~0.05 in the two cases. These results verify our first hypothesis. They show that the QE behavior occurs more frequently in successful tasks than unsuccessful ones. In addition, the results show that QE duration is longer in successful tasks than unsuccessful ones.

\subsection{Robotic Surgery Data Set Results}
\label{subsec:QE_Robotic Surgery Data Set Results}

The results of the robotic surgery data set are used to verify our second hypothesis. The results show that 40\% of the more experienced surgeon's videos have QE, compared with 0\% of the less experienced surgeon's videos. The average QE duration in the more experienced surgeon's videos is 867~$\pm$~480 ms, compared with zero ms in the less experienced surgeon's videos, since QE was not found in any of them. 

We conducted two hypothesis tests on the above results. A Chi-square test (with $\alpha$ = 0.05) was used to assess whether the QE phenomenon occurs more often with the more experienced surgeon than with the less experienced one. The test was statistically significant with ${\chi}$$^2$(1) = 6.864 and \textit{p}~$<$~0.05. Moreover, a Mann–Whitney U test was used to assess whether QE duration is longer in the more experienced surgeon's case than the less experienced one. This test was also statistically significant with \textit{p}~$<$~0.05. These results verify our second hypothesis.  

\section{Discussion}
\label{sec:Discussion}

In this work, we report the discovery of the QE behavior in two independently collected data sets in two MIS settings. Our results show that the QE behavior is more likely to be observed in successful targeting tasks (compared with unsuccessful ones) and in targeting tasks performed by more experienced surgeons (compared with less experienced ones). In addition, the results show that QE duration is significantly longer in these two cases. Besides, our results show that the QE duration is potentially a stronger distinguishing factor between different performances (based on the success of the performance or the experience level of the surgeon) than the existence of the QE behavior on its own. These results are consistent with previous findings in the sports training literature~\cite{vickers2007perception}. 

One limitation, though, of this study is the small number of surgeons whose data were used in our investigation. We used data from 20 surgeons in the first data set and from two surgeons in the second data set. More data sets, with more surgeons and more targeting tasks, are needed to reach more conclusive results and investigate the QE behavior in a variety of other targeting tasks in MIS. 

Nevertheless, the potential impact of this study spans the areas of training, skill assessment and human-robot interaction in MIS. For training, this work opens the door to testing QE training~\cite{digmann2018quiet} in MIS settings, with the evidence it provides that QE behavior does exist in this setting. Since the characteristics of the QE behavior are well defined, we think that it can be relatively easy to integrate QE training into existing surgical training methods. Furthermore, QE training has been shown to be robust against high levels of anxiety and pressure~\cite{vine2014quiet}, which can make it even more valuable for surgical training in MIS. 

The results of our study can be useful for surgical skill assessment in MIS. Learning how the QE behavior changes as the surgeon's skill/experience level increases can open the door to monitor these changes and use them as a tool to classify the skill levels of practicing surgeons based on their QE behavior. Furthermore, it would be interesting to investigate the effect of adding the QE behavior related features into existing machine learning based skill assessment systems~\cite{vedula2017objective} on their classification performance. Moreover, the tasks we considered in this study provide a clue of the family of the surgical tasks where using the QE behavior can be useful for skill assessment.

Our study can lead to some important implications on the design of human-robot interfaces in RAS. In particular, our study highlights the importance of collecting and investigating eye gaze data in this setting. This provides additional evidence of the potential value of integrating eye gaze tracking in new and existing RAS platforms and simulation environments~\cite{tong2015retrofit}~\cite{nathan2021senhance}. 
 
\section{Conclusion and Future Work}
\label{sec:Conclusion and Future Work}

In this paper, we report our discovery of the QE phenomenon in MIS. We investigated the existence and duration of the QE behavior in two independently collected data sets in a sinus surgery setting and a robotic surgery setting. In two targeting tasks, we observed that the QE behavior occurs more frequently and for longer durations in successful task performances (compared with unsuccessful ones) and in performances by experienced surgeons (compared with less experienced ones). These observations are consistent with similar findings in sports training, where the QE behavior has been successfully used to improve the motor skills of novice trainees. 

This work opens up several new lines of research in surgical training and skill assessment in MIS. For instance, more investigations are needed to identify more tasks where the QE behavior would occur. These tasks can then be used to test the effect of integrating QE training into the traditional training approaches of these tasks. The integration of the QE training has the potential to improve the motor skill learning aspects of these tasks and can potentially make the traditional training methods more robust to situations of high anxiety and pressure. Furthermore, it would be interesting to include the well-defined characteristics of the QE behavior as novel features in skill assessment machine learning models and evaluate how they can improve their performance.

%add a cknoweldgement section for Tony Jarc regarding the data set if we are not having him as a co-author

\section*{ACKNOWLEDGMENT}

We would like to thank Maram Sakr and Tim Powers for their assistance with the statistical analysis of the results of this work.
 
\bibliographystyle{IEEEtran}  
\bibliography{biblio}    % refs.bib --> should contain all references

\vfill

\end{document}